\documentclass{article}
\usepackage{ijcai16}

\usepackage{times}
\usepackage{graphicx}
\usepackage{amsmath}
\usepackage{amssymb}
\usepackage{array}
\usepackage{booktabs}
\usepackage{hyperref}

\graphicspath{{.}}
\pdfoutput=1

\title{The Role of Typicality in Object Classification: \\ Improving The Generalization Capacity of Convolutional Neural Networks}
	\author{\href{mailto:babaks@cs.rutgers.edu}{Babak Saleh}\\
		\small Department of Computer Science\\ 
		\small Rutgers University\\
		\And
	    	    \href{mailto:elgammal@cs.rutgers.edu}{Ahmed Elgammal}\\
		\small Department of Computer Science \\ 
		\small Rutgers University\\
		\And
				\href{mailto:jacob@ruccs.rutgers.edu}{Jacob Feldman} \\
		\small Center for Cognitive Science \\ 
		\small Rutgers University}

\begin{document}

\maketitle

\begin{abstract}
	Deep artificial neural networks have made remarkable progress in different tasks in the field of computer vision. However, the empirical analysis of these models and investigation of their failure cases has received attention recently. In this work, we show that deep learning models cannot generalize to atypical images that are substantially different from training images. This is in contrast to the superior generalization ability of the visual system in the human brain. We focus on Convolutional Neural Networks (CNN) as the state-of-the-art models in object recognition and classification; investigate this problem in more detail, and hypothesize that training CNN models suffer from unstructured loss minimization. We propose computational models to improve the generalization capacity of CNNs by considering how typical a training image looks like. By conducting an extensive set of experiments we show that involving a typicality measure can improve the classification results on a new set of images by a large margin. More importantly, this significant improvement is achieved without fine-tuning the CNN model on the target image set. 
	%
	%
	
\end{abstract}

\section{Introduction}
\label{sec:intro}
Convolutional Neural Networks (CNN) have made remarkable progress in a variety of computer vision tasks. To just name few of the recent advances, CNN-based models greatly improved the performance in object classification~\cite{simonyan2014very}, object detection~\cite{ren2015faster}, image retrieval~\cite{sharif2015baseline}, fine-grained recognition~\cite{lin2015bilinear}, scene classification~\cite{zhou2014object}, action classification~\cite{rstarcnn} and generating image descriptions ~\cite{vinyals2014show}.
%
\begin{table}
	\begin{center}
		\resizebox{.5\textwidth}{!}{
			\begin{tabular}{|c|c|c|c|c|c|c|}
				\hline
				Method & \multicolumn{3}{c|}{Top-1 error ($\%$)} & \multicolumn{3}{c|}{Top-5 error ($\%$)} \\
				\cline{2-7}
				& Train & Test-T & Test-A & Train & Test-T& Test-A \\ 
				\hline \hline
				AlexNet~\cite{krizhevsky2012imagenet}  & 38.1 & 49.5 & 74.96 & 15.32 & 24.01 & 47.07 \\
				OverFeat~\cite{sermanet2013overfeat}& 35.1 & 45.36 & 75.62 & 14.2 & 22.27& 46.73 \\
				Caffe~\cite{jia2014caffe}& 39.4 & 51.88 & 77.12 & 16.6 & 24.74 & 46.86 \\
				VGG-16~\cite{simonyan2014very} & 30.9 & 44.04 & 77.82  & 15.3 & 26.31& 47.49  \\
				VGG-19~\cite{simonyan2014very} & 30.5 & 43.72 & 76.35 & 15.2 & 26.85 & 45.99 \\
				\hline
			\end{tabular}}
		\end{center}
		\vspace*{-10pt}
		\caption{\textit{State-of-the-art Convolutional Neural Networks (trained on normal images) fail to generalize to atypical/abnormal images for the task of object classification. Columns ``Train" show the reported errors on typical/normal images (ILSVRC 2012 validation data), while numbers in the next two columns are the errors on our atypical ``Test-A", and typical ``Test-T" images. The significant drops in performance, especially when test on atypical images, show the limited generalization capacity of CNN. Our goal is to enhance  these visual classifiers and reducing this gap.}}
		\label{tab:deeplearning}
		\vspace*{-5pt}
\end{table}

\begin{figure*}
	\begin{center}
	\label{fig:intro_cls}
	\includegraphics[width=2\columnwidth, height=.5\columnwidth]{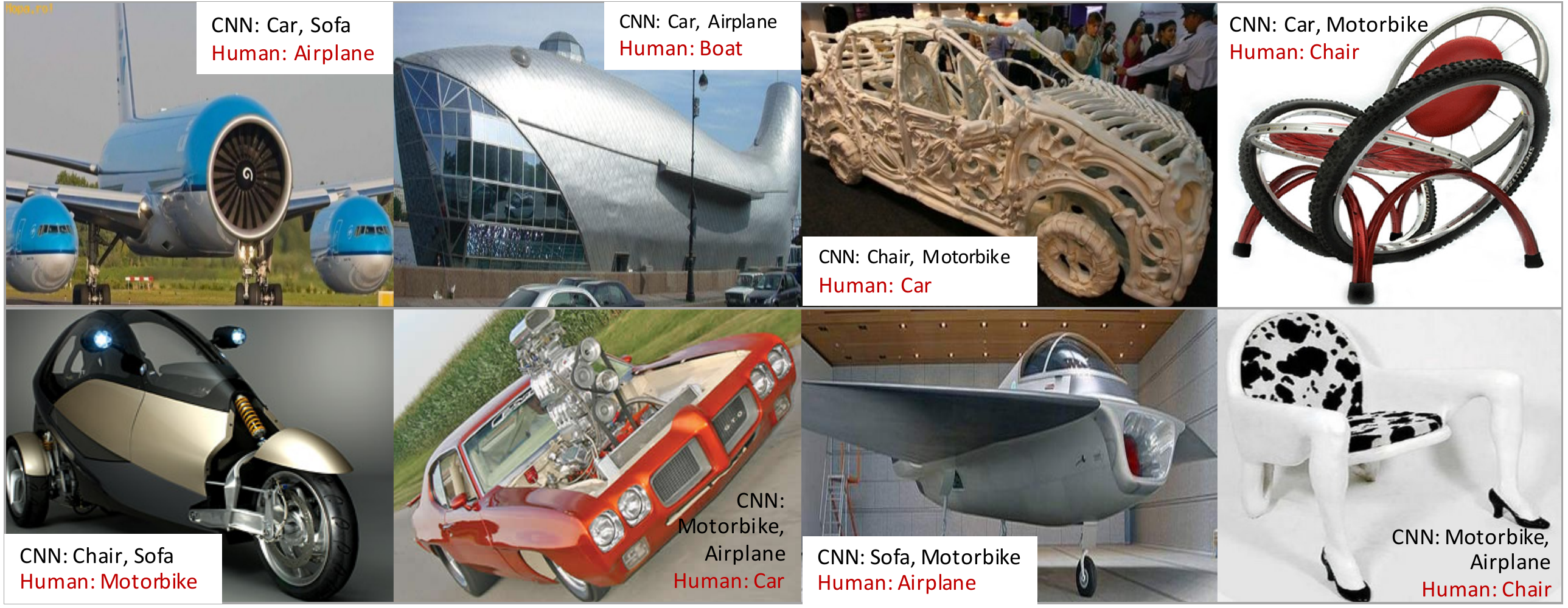}
	\caption{\textit{Some atypical images that are misclassified by a fine-tuned CNN, where as humans can categorize them correctly. Top two model predictions (in black) are reported, where the first one has 100 \% model confidence.}}
\vspace*{-8pt}
\end{center}
\end{figure*}

Despite surpassing the human categorization performance on large-scale visual object datasets~\cite{he2015delving}, convolution neural networks cannot emulate the generalization power of the human visual system in real-world object categorization ~\cite{ghodrati2014feedforward,pinto2008real}, especially when it comes to objects that differ substantially from the training examples. Figure~\ref{fig:intro_cls} shows examples of these atypical images, which human subjects categorize correctly, but which a CNN model misclassified with a high confidence. We evaluate the performance of state-of-the-art CNNs for the purpose of object classification on atypical images. Humans are capable of perceiving atypical objects and reasoning about them, even though they had not seen them before~\cite{saleh2013object}. But our experiments have shown that state-of-the-art CNNs failed drastically to recognize atypical objects. Table~\ref{tab:deeplearning} shows the results of this experiment, where we took off-the-shelf CNNs and applied them on atypical images. The significant performance drop, when tested on atypical images, is rooted in the limited generalization power of CNN models versus the human visual system.

%
	
One might argue that this issue of cross-dataset generalization is implicitly rooted in dataset biases, and not limited to CNN models~\cite{torralba2011unbiased}. However, we argue that the huge number of labeled images in the training set of these models (here ImageNet) should alleviate this drawback. By providing a wide range of variation in terms of visual appearances of objects in training images, the effect of biases fades away. We support our argument by testing same networks on a new set of images that are disjoint from the training set of ImageNet~\cite{deng2009imagenet}, but look typical. Results of this experiment as it is reported in columns ``Test-T" in Table~\ref{tab:deeplearning} show a much smaller drop in accuracy, compared to the case of testing on atypical images (Test-A). We conclude that dataset bias can affect the performance of CNNs for object categorization, but it is not the main reason behind its  poor generalization to new datasets.
	

Instead, inspired by the way humans learn object categories, we can empower CNN models with the ability to categorize extremely difficult cases of atypical images. Humans begin to form categories and abstractions at an early age\cite{bigbook}. The mechanisms underlying human category formation are the subject of many competing accounts, including those based on prototypes\cite{minda2001prototypes}, exemplars\cite{nosofsky1984choice}, density estimation\cite{ashby1995categorization}, and Bayesian inference\cite{goodman2008rational}. But all modern models agree that human category representations involve subjective variations in the typicality or probability of objects within categories. In other words, typicality is a graded concept and there is no simple decision boundary between typical vs. atypical examples. A category like bird, would include both highly typical examples such as robins, as well as extremely atypical examples like penguins and ostriches, which while belonging to the category seem like subjectively ``atypical" examples.  Visual images can also seem atypical, in that they exhibit features that depart in some way from what is typical for the categories to which they belong. Humans learn object categories and form their visual biases by looking at typical samples ~\cite{Sloman,Rips}. But they are able to recognize atypical/abnormal objects which show significant visual variations from training set, without even observing them at the learning stage. 
%

From computer vision and machine learning perspectives, state-of-the-art object classification and detection is based on discriminative models (e.g. SVM, CNN, Boosting) rather than generative ones. Discriminative training focuses more on learning boundaries between object classes, instead of finding common characteristics in each class. Training CNN models is based on minimization of a loss function, defined as the misclassification of training samples. In that sense CNN implicitly emphasizes on the boundary between classes. However, training samples are not weighted based on how typical they look like, or equivalently how representative they are for a given category.
 
In this work, we hypothesize that not all images are equally important for the purpose of training visual classifiers, and in particular deep convolutional neural networks. Instead, we show that if training images are weighted based on how typical they look, we can learn visual classifiers with a better generalization capacity. Our final CNN model is fine-tuned only with typical images, but outperforms the baseline model (training samples are not weighted) on dataset of atypical images. We also empirically compare a large set of functions that can be used for weighting samples, and conclude that an even-degree polynomial function of typicality ratings is the best strategy to weight training images. We also investigate the effect of loss functions and depth of network by conducting experiments on two datasets of ImageNet and PASCAL. 

The main contributions of this paper are as following: 
	\begin{itemize}
		\item Evaluating CNN models on datasets of images that are different from training data, and characterizing failure cases as the poor generalization capacity of CNN models. Especially contrasting these failures to the superior performance of humans in categorizing atypical objects.
		\item Inspired by theories in psychology and machine learning, we propose three hypotheses to improve the generalization capacity of CNN models. These hypotheses are based on weighting train images depending on how typical they look. 
		\item We conduct an extensive set of experiments, to empirically compare different functions of typicality rating for weighting training images.
	\end{itemize}

\section{Related Work}
\label{sec:rel}
Space does not allow an encyclopedic review of the prior literature on deep learning, but we refer interested readers to literature reviews of~\cite{lecun2015deep}. For our research, we focus on convolutional neural networks~\cite{fukushima2013artificial,krizhevsky2012imagenet,lecun1998gradient} as the state-of-the-art deep learning model for the task of object recognition. CNN~\cite{lecun1989backpropagation} has its roots in Neocognitron~\cite{fukushima1980neocognitron}, which is a hierarchical model based on the classic notion of simple and complex cells in visual neuroscience~\cite{hubel1962receptive}. However, CNN has additional hidden layers to model more complex non-linearities in visual data and its overall architecture is reminiscent of the LGN $\mapsto$ V1  $\mapsto$ V2 $\mapsto$ V4 $\mapsto$ IT hierarchy in the visual cortex ventral pathway. Additionally it uses an end-to-end supervised learning algorithm, called ``Backpropagation" to learn weights of layers.  Different variations of CNN models have made breakthrough performance improvements in a variety of tasks in the field of computer vision.

Despite an extensive amount of prior works on application of CNN and proposed variations of it, theoretical understanding of them remains limited. More importantly, even when CNN models achieve human-level performance on visual recognition tasks~\cite{he2015delving}, what will be the difference between computer and human vision? On the one hand, Szegedy et al.\shortcite{szegedy2013intriguing} demonstrated that CNN classification can be severely altered by very small changes to images, leading to radically different CNN classification of images that are indistinguishable to the human visual system. On the other hand, Nguyen et al.\shortcite{nguyen2015deep} generated images that are completely unrecognizable by humans, but which a CNN model would classify them with 99.99\% confidence. This strategy to fool CNN models, raises questions about the true generalization capabilities of such models, which we investigate it in this paper. 

In addition, recent studies in the field of neuroscience and cognition have shown the connection between deep neural networks (mainly CNN) and the visual system in human brain. Yamins et al.\shortcite{yamins2014performance} have shown correlation (similarity) between activation of middle layers of CNN and brain responses in both V4 and inferior temporal (IT), the top two layers of the ventral visual hierarchy. Cadieu et al.\shortcite{cadieu2014deep} proposed a kernel analysis approach to show that deep neural networks rival the representational performance of IT cortex on visual recognition tasks. Khaligh-Razavi and Kriegeskorte\shortcite{khaligh2014deep} studied 37 computational model representations and found that the CNN model of~\cite{krizhevsky2012imagenet} came the closest to explaining the brain representation. Interestingly, the amount of correlation between human IT and layers of CNN increases by moving to higher layers (fully-connected layers). They concluded that weighted combination of features of the last fully connected layer can explain IT to a full extent. Also it has been shown that CNN models predict human brain activity accurately in early and intermediate stages of visual pathway~\cite{agrawal2014pixels}.
%
 
 \begin{figure}
 	\begin{center}
 	\includegraphics[width=.35\paperwidth, height = .18\paperwidth]{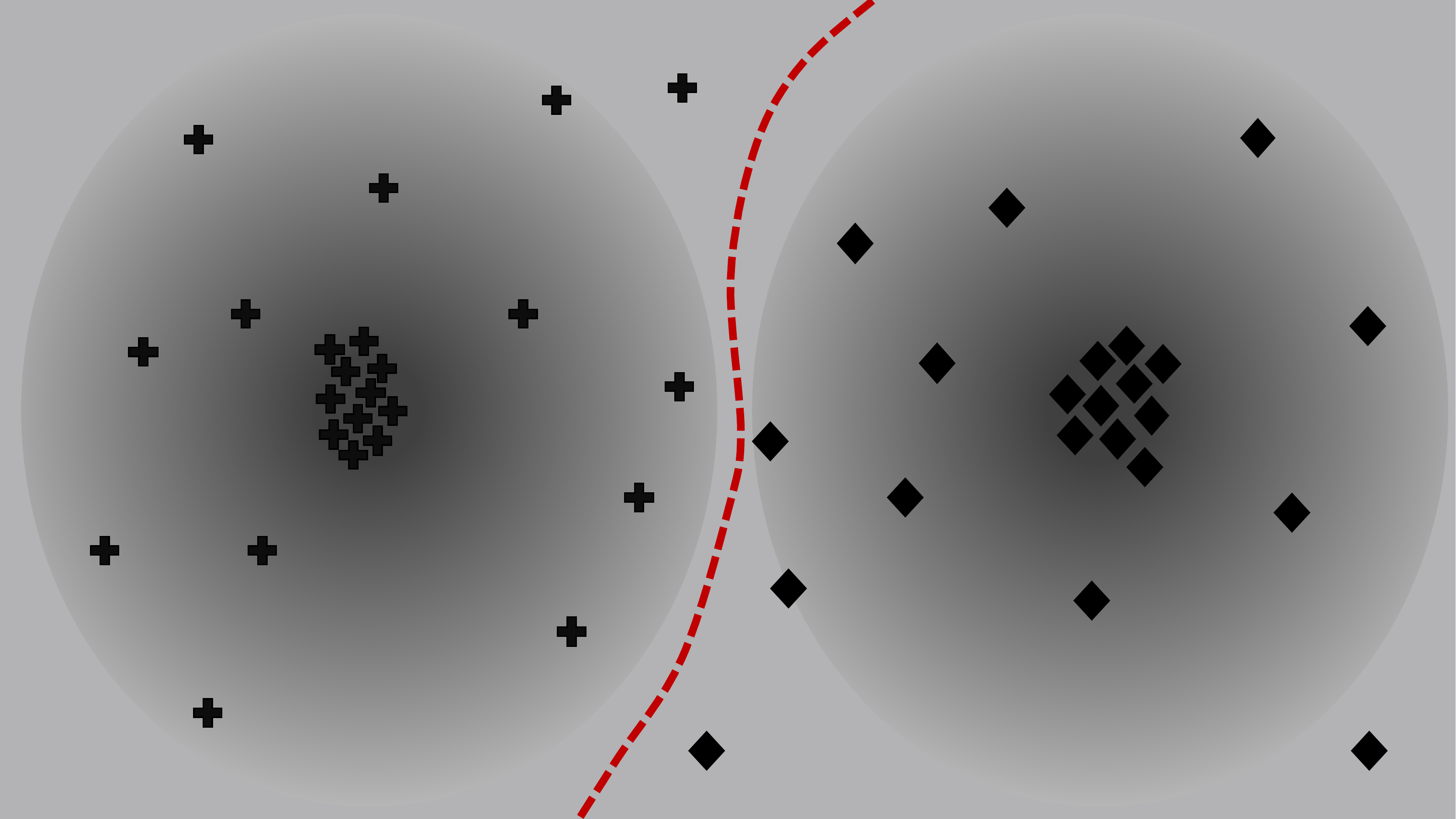}
 	\vspace*{-5pt}
 	\caption{\textit{Illustration of the notion of typicality. Examples of two classes of cross and diamond show different shades (degrees) of typicality. While we can find the red classifier to discriminate classes, we cannot find a decision boundary between atypical vs. typical samples of the category of interest.}}
 	\label{fig:model}
 	\vspace*{-10pt}
 \end{center}
 \end{figure}

There are some prior works on finding the right features~\cite{blum1997selection}, choosing the appropriate train set and how to order training examples for learning better classifiers~\cite{bengio2009curriculum}. Also, It has been shown that CNN models benefit from training with larger datasets of images. This is because the greatest gains in detection performance will continue to derive from improved representations and learning algorithms that can make efficient use of large datasets~\cite{zhu2015we}. However, this leaves open the question of how the visual classifier should weight all training images, equally or not?

\section{Computational Framework}
\label{sec:model}
In this section we first go through theoretical background and compelling theories about learning visual concepts in both fields of psychology and computer vision. We explain how one can measure typicality of objects in an image. Then we propose three hypotheses to use these typicality scores to improve generalization capacity of visual classifiers. 

\subsection{Framework Motivation}

Humans learn a visual object class by looking at examples that are more representative for that object category, or what is called typical samples~\cite{Sloman,Rips}. It has been shown that children who learn a category by looking at more typical samples, later can recognize its members better~\cite{Rosch_1978}. If training examples look more typical, they fall close to each other in an underlying space of visual features. This learning strategy not only helps humans to form a concept, but also allows them to more easily apply the concept to novel images. This great ability of human visual system allows them to recognize completely different variations of an object, even to the extent of atypical objects. This suggests that emphasizing typical examples might be helpful in improving the generalization ability of classifiers.


However, state-of-the-art object classifiers in computer vision are discriminative models, where they distinguish different objects by learning category boundaries. CNN models as discriminative deep neural networks have multiple layers to learn a hierarchy of visual features and categorize objects based on loss minimization as a function of misclassification errors. In other words, if an image is classified correctly (usually the case for typical images), it has little or no impact on the loss function, hence can be ignored in training. This implies that examples close to the decision boundary, which are likely to be more atypical images, play a substantial role in learning CNN models. This suggests training CNN emphasize on more atypical images to learn visual classifiers with a better performance. 
Figure~\ref{fig:model} illustrates the concept of typicality where examples of two classes ($\mathcal{C}$) are shown with diamonds and crosses, and the red dotted line is one possible decision boundary. There are two main points to be taken from this illustration:

\textbf{First}, as we discussed in Section\ref{sec:intro} typicality is a graded concept, which directly relates to the likelihood of an observation given its class distribution $\mathcal{P}(\mathcal{X}|\mathcal{C})$. Very typical examples are expected to be located close to the mean of each class distribution (center of clouds), with a high probability~\cite{feldman2000bias}. Moreover, as we move away form the center, we still observe examples of the same category. But every member of the category shows a different rate of typicality  $\mathcal{P}(\mathcal{X}|\mathcal{C})$. This is visualized as a smooth transition when moving away from the center of a class. More importantly there is no clear boundary between typical and atypical members.

\textbf{Second}, atypicality happens for a variety of reasons. This is visualized as there is not a unique axis for transition from darker to brighter shades of gray. Although examples close to the decision boundary might be atypical for their category; but the atypical examples are more diverse and not limited to the boundary examples. In conclusion, the two sets of atypical and boundary examples are not equal. 

%

\subsection{Sample-Based Weighted Loss}
CNN architecture consists of multiple blocks, where each block has convolution layer followed by pooling and normalization layers. On the top of these blocks, there are fully connected layers that are designed to learn more complex structures of object categories. The last layer of CNN computes the ``loss" as a function of mismatch between model prediction and ground truth label. The training of CNN is formulated as minimization of this loss function~\cite{lecun1989backpropagation}. However, our work is the first study to analyze the effect of weighting samples and using different loss functions incorporating in typicality scores, for improving generalization of CNN. We associate each sample $X$ with a weight $\tau$ as a function of its typicality, which we explain later. We build our models by weighting samples based on two loss functions: \textit{Softmax log} and \textit{Multi-class structured hinge}. While the first one is the fastest and widely used in prior works, the later is more suitable for our purpose.


\paragraph{Softmax log loss: } For classification problems using deep learning techniques, it is common to use the softmax of one of  the $\mathcal{C}$ encodings at the top layer of the network, where $\mathcal{C}$  is the number of classes. Assuming the output to the \textit{i-th} node in the last layer, for the image $\mathcal{X}$ is: $z_{i}(\mathcal{X})$.  Then our goal is to minimize the weighted multinomial logistic loss ($\mathcal{L}$) of its softmax over N training images : 
\begin{gather*}
\mathcal{L} = \sum_{n } - \tau(\mathcal{X}_{n})*  \log (\sigma_{i}(\mathcal{X}_{n})) \hspace*{10pt} (n = 1,...,N )
\\
\sigma_{i}(\mathcal{X}_{n}) = \exp(z_{i}(\mathcal{X}_{n})) / \sum_{j}\exp(z_{j}(\mathcal{X}_{n})), (i,j = 1,...,\mathcal{C}).
\end{gather*}

\paragraph{Multi-class structured hinge loss: } It is also known as the Crammer-Singh loss, and widely used for the problem of structured prediction. This loss function is similar to hinge-loss, but it is computed based on the margin between score of the desired category and all other prediction scores ( $\phi(i)$)~\cite{crammer2002algorithmic}. We aggregate this loss function ($\mathcal{L}$) by a  weighted summation over training samples:
\begin{gather*}
\mathcal{L} = \sum_{n}  \tau(\mathcal{X}_{n}) * \max(0, 1-\phi_{i}(\mathcal{X}_{n})) 
\\
\phi_{i}(\mathcal{X}_{n}) = z_{i}(\mathcal{X}_{n}) - max_{i \neq j}(z_{j}(\mathcal{X}_{n})).
\end{gather*}

Multi-class hinge loss is particularly of our interest as it considers the margin between all class predictions. This is an importance piece of information when we want to generalize our visual classifiers to the case of atypical objects. These examples are harder to be classified, and as a result the class prediction is not a distribution with its peak around the desired class. In fact, the object might get high class confidence for multiple categories. This results in a smaller $\phi$ and bigger $\mathcal{L}$. Consequently, this loss would implicitly favor atypical examples as they generate larger losses.

\subsection{Measuring Typicality of Objects}
\label{subsec:model_typ}
We have two approaches for measuring the typicality of objects. On the one hand, we compute the probability score $\mathcal{P}(T | \mathcal{X})$ as how typical $(T)$ is the object only based on its visual features $\mathcal{X}$. For the case of class-specific typicality we can infer: $\mathcal{P}(T | \mathcal{X}) \propto \mathcal{P}(X | \mathcal{C})$ where $\mathcal{C}$ indicates the category, and independent of the class: $\mathcal{P}(T | \mathcal{X}) \propto \mathcal{P}(\mathcal{X} )$. Then its complement ($ 1 - \mathcal{P}(T | \mathcal{X})$) is the probability of atypicality.

To implement this probability, we use one-class SVM where only positive samples of one category (here typical images) are used and there is no negative (atypical) training example. This model can be understood as a density estimation model where there is no prior knowledge about the family of the underlying distribution. We learn this one-class SVM in two scenarios: 1) General class-independent typicality: all images are used; 2) Class-specific typicality: for each category one SVM is trained only based on typical images of the category of interest. We refer to these models as ``external score of typicality". This is because these scores are computed using a model distinct from CNN, and based on visual features different from what we use for object categorization. These scores are computed offline for all training images and not changing over different epochs of CNN training.

On the other hand, we can judge typicality of training images directly from the output of CNN visual classifiers. Lake et al.\shortcite{lakedeep} showed that the output of the last layer of CNN models can be used as a signal for how typical an input image looks like. In other words, typicality ratings are proportional to the strength of the classification response to the category of interest. Assuming loss is defined over $C$ object categories and there are $N$ nodes in the last layer, we compute ``internal probability of typicality" as:
\begin{eqnarray}
Z_{i} = \exp(y_{i}) / \sum_{j=1}^{C}\exp(y_{j}) ;  \hspace*{7pt}   where:  y_{j} = \sum_{i=1}^{N}x_{i}W_{ij}
\end{eqnarray}
Alternatively, we use the entropy of a category prediction as a measure of uncertainty in responses, which punishes more uncertain classifications. We call this ``internal entropy of typicality" and compute it as : $-Z_{i}log(Z_{i})$.

\subsection{Hypotheses}
\label{subsec:hyp}
We propose three hypotheses to improve generalization of visual classifiers, especially when the test image looks substantially different from training images (atypical). 

\textbf{First, } Inspired by the prototype theories from psychology, we hypothesize that learning with more emphasis towards representative  (typical) samples would increase the generalization capacity of the visual classifier.

\textbf{Second, } Learning with emphasis on more atypical examples in the training set would enhance the generalization capacity. This is  because it complements the way that loss function emphasizes boundary examples. This hypothesis, places additional emphasis on other possible directions of atypicality in training data that might not be on the boundary. 

\textbf{Third, } We hypothesize that emphasizing both typical and atypical examples might be the key for a better generalization performance, and should be used for learning visual classifiers. The main idea behind this hypothesis is the fact that any visual classifier should learn how the object category is formed (mainly typical examples), and how much a variation it would allow for its members (atypical samples). 

To implement the first two hypotheses we multiply the loss of each sample by $\tau(\mathcal{X})$, which is a function of typicality (for the first hypothesis) or atypicality (second hypothesis). 
To investigate the effect of different functions of typicality score, we evaluate exponential ($\exp{\mathcal{P}(T | \mathcal{X}) }$) and gamma ($\gamma^{\mathcal{P}(T | \mathcal{X}) }$) functions to emphasize typicality versus a logarithmic function ($-\log(\mathcal{P}(T | \mathcal{X}) )$) to emphasize atypicality. This helps us to evaluate the generalization capacity of a CNN model, when trained with non-linear weighting. We evaluate our last hypothesis by implementing the weighting function as an even-degree polynomial:
\begin{eqnarray}
\mathcal{F}(T) = \alpha  (T - \mu)^d  + \beta; \hspace*{10pt} d = 2k (k=1,...,n)
\end{eqnarray}
These functions are symmetric around the average typicality score in the dataset ($\mu$), and place emphasis on data points in both extremes of the typicality axis.

\vspace*{-10pt}
\begin{table}[t]
	\begin{center}
		\scalebox{.8}{
			\begin{tabular}{|c|c|c|c|c|c|}
				\hline
				Loss & Test set  & Typ & Atyp & cls-typ & cls-atyp \\
				\hline \hline
				MS-Hinge & Atypical& 68.58  & 70.64 & 70.84 &  68.47\\
				Softmax & Atypical& 63.69  & 66.82 & 65.81 & 66.48\\
				\hline
				MS-Hinge & Typical  & 79.90 & 84.07 & 82.88 &   83.40  \\
				Softmax & Typical& 77.11 & 80.42 & 83.40 & 82.96\\
				\hline
			\end{tabular}}
		\end{center}
		\vspace*{-12pt}
		\caption{\textit{Object classification accuracy (\%) of the AlexNet on two test sets of Typical(lower box) and Atypical(upper box) images. Two loss functions (rows) are compared, when training samples are weighted via four functions (columns): Typicality (first), Atypicality (second), Class-specific typicality (third) and Class-specific atypicality (fourth).}}
		\vspace*{-5pt}
		\label{tab:loss}
	\end{table}
%

\section{Experimental Results}
\label{sec:exp}
\subparagraph{Datasets:}
We used three image datasets: 1) ImageNet challenge (ILSVRC 2012 \& 2015), 2) Abnormal Object Dataset~\cite{saleh2013object}, 3) PASCAL VOC 2011 train and validation set. 
We conducted our experiments with six object categories: Aeroplane, Boat, Car, Chair, Motorbike and Sofa. We did this to be able to verify our generalization enhancement for atypical images in Abnormal Objects dataset, which contains these categories. We merged related synsets of ILSVRC 2012 to get images of these categories, resulting in 16153 images, which we refer to as ``train set I".

Additionally, we experimented with train and validation set of PASCAL 2011. This is needed because due to a higher level of supervision in PASCAL data collection process, images are more likely to look typical. However, ImageNet data shows significant variations in terms of visual appearance (pose, missing or occluded parts, etc.) that can make the image and object look less typical. We collected 4950 images from PASCAL dataset, which we refer to as ``train set II".
We also used a subset of 8570 images from ILSVRC 2015 detection challenge, which we call ``test typical", and are completely disjoint from the set used in training. Images of~\cite{saleh2013object} form our ``test atypical" set, which represents confirmed atypical/abnormal objects. 

\begin{table}[t]
	\begin{center}
		\scalebox{0.75}{
			\begin{tabular}{|c|c|c|c|c|}
				\hline
				Weighting & \multicolumn{4}{c|}{Mean Accuracy (\%)}\\
				\cline{2-5}
				Function in & \multicolumn{2}{c|}{Test Atypical} & \multicolumn{2}{c|}{Test Typical}\\
				\cline{2-5}
				Fine-Tuning & Epoch 1 & Epoch 10 & Epoch 1 & Epoch 10 \\
				\hline \hline
				No weight & 56.39  & 65.18& 78.15 & 83.51 \\
				Random&  57.15 & 66.45&  73.60 &  83.84\\
				\hline
				Typicality &  64.53 & 68.58 & 69.22 & 79.90  \\
				Atypicality& 66.61 & 70.65  & 75.82  &  84.07\\
				\hline
				CLS-Typ&   67.25 & 70.84&  77 & 81.88  \\
				CLS- Atyp&  63.26 & 68.46 &  76.96  &  83.40 \\
				\hline
				Log-Typ &   64.38 & 68.28 & 78.80 &  83.67 \\
				Log CLS-Atyp&  64.21 & 67.80 & 76.13  &  83.24  \\
				Memorability&  64.69  & 68.33 & 76.31 &  83.96 \\
				\hline			
				Poly Deg-2&  59.13  &69.49  & 80.03 & 84.42 \\
				Poly Deg-4& 60.22  & 71.52 &  77.74 & 83.45 \\
				Poly Deg-6&  60.86 &70.31  & 77.66 & 84.22 \\
				\hline			
				In-Probability&  65.97 & 69.53  & 80.71 & 85.82 \\
				In-Entropy& 60.54 & 68.05   &  79.44 & 82.29 \\
				In-Prob +Atyp&  62.94 & 68.21  & 75.82 & 83.09 \\
				\hline			
			\end{tabular}}
		\end{center}
		\vspace*{-10pt}
		\caption{\textit{Performance of object classification with AlexNet fine-tuned on ``Train Set I", when MS-Hinge loss is used. Rows show different functions of typicality scores.}}
		\vspace*{-10pt}
		\label{tab:scores_imagenet}
	\end{table}
	\vspace*{-8pt}
	
\subparagraph{Typicality estimation:}
We measured the typicality of images via one-class SVMs in two settings: General and Class-specific. The first case is independent of the object-category and only measures how typical the input image looks in general. But, for the latter we trained six (one for each category) one-class SVMs with typical images of the category of interest. We extracted kernel descriptors of~\cite{bo2010kernel} at three scales as the input features.

\subparagraph{Visual classifier:}
We evaluated our three hypotheses with the CNN model of AlexNet~\cite{krizhevsky2012imagenet}. 
Nevertheless, our approach can be used for other state-of-the-art CNN models of object classification as well. We acquired the implementation of Caffe~\cite{jia2014caffe} for AlexNet and fine-tuned the network for all the following experiments. 
\subsection{Comparison of Loss Functions}
In order to find the proper loss function in fine-tuning the network, we conducted an experiment with two losses: Softmax and Multi-structured hinge (MS-Hinge) loss. For this experiment we only fine-tuned the last fully-connected layer with ``Train Set I". Table~\ref{tab:loss} shows the performance comparison based on using different loss functions and weighting methods. We conclude that independent of the weighting strategy, Multi-structured hinge loss (MS-Hinge) performs better than the Softmax loss. Consequently, the rest of experiments were conducted based on fine-tuning with MS-hinge loss.



\subsection{Comparison of Weighting Functions}
We conducted a set of experiments to compare the performance of CNN models for the task of object classification, when fine-tuned using different weighting functions. Table~\ref{tab:scores_imagenet} shows the result of these experiments on the two test sets of Typical and Atypical. We report the mean accuracy after the first and tenth epochs. While the result of the first epoch indicates how fast the network can learn a category, the tenth epoch elaborates the performance when the network has matured (trained for a longer time).

%

\subparagraph{External score of typicality:}
The first box in Table~\ref{tab:scores_imagenet} shows baseline experiments, when the first row is the AlexNet fine-tuned without any weighting. Second row shows weighting training images with a random number between zero and one. The result shows that randomly weighting training data does not have any effect on improving the generalization performance of the trained network.
Next box shows the results of using the typicality and atypicality probabilities. We conclude fine-tuning with raw atypicality/typicality weighting can significantly enhance the generalization of CNN, even after the first epoch. However, fine-tuning with raw typicality can degrade the performance, when tested on typical images. The third box has similar results, where typicality or atypicality are computed based on class-specific one-class SVMs.
%

Fourth box in Table~\ref{tab:scores_imagenet} investigates the importance of nonlinear weighting functions. First and second row are the results of using logarithmic functions, where $\tau(\mathcal)$ is either typicality score (first row) or class-specific atypicality scores (second row). We conclude that networks do not gain much from non-linear functions of either typicality or atypicality scores, when test on atypical images. But non-linearities help stabilizing the performance on typical images. The last row of the fourth box, indicates that fine-tuning AlexNet with the memorability score~\cite{ICCV15_Khosla} will increase its generalization performance (comparing to baselines). However, fine-tuning with memorability do not outperform typicality weightings. 
The fifth box in Table~\ref{tab:scores_imagenet} evaluates our third hypothesis, where three polynomials are used for weighting the training samples. In general, this strategy outperforms other methods (comparing the tenth epoch performance) on atypical test set, and do not degrade the performance on typical set (compare to baseline).

%
\begin{table}
	\begin{center}
		\scalebox{0.75}{
			\begin{tabular}{|c|c|c|c|c|}
				\hline
				Weighting  & \multicolumn{4}{c|}{Mean Accuracy (\%)}\\
				\cline{2-5}
				Function in& \multicolumn{2}{c|}{Test Atypical} & \multicolumn{2}{c|}{Test Typical}\\
				\cline{2-5}
				Fine-Tuning& Epoch 1 & Epoch10 & Epoch 1 & Epoch10 \\
				\hline \hline
				No weight & 30.03  & 48.40 &51.22  & 64.17  \\
				Random& 29.22 & 49.18& 49.03 &  58.9\\
				\hline
				Atypicality & 41.21 & 52.24  &  55.3 & 70.28  \\
				Log-Typ & 36.95 &  50.80  & 52.76 & 68.88 \\
				Exp-Typ & 41.95 & 53.79  & 54.31& 67.51  \\
				Memorability&  40.09 & 49.36 & 52.83 & 69.6 \\
				\hline			
				Poly Deg-2&   41.37 &  55.44 & 54.8 & 73.02 \\
				Poly Deg-4&  42.33 & 56.39  &  53.9 & 72.42 \\
				Poly Deg-6&  44.73  &  52.72 & 52.93 &  72.7\\
				\hline			
			\end{tabular}}
		\end{center}
		\vspace*{-10pt}
		\caption{\textit{Performance on object classification with AlexNet fine-tuned on ``Train Set II", when MS-Hinge loss is used. Rows show different functions of typicality scores.}}
		\vspace*{-10pt}
		\label{tab:scores_PASCAL}
	\end{table}

\vspace*{-15pt}
 \subparagraph{Internal score of typicality: }
The last box in Table~\ref{tab:scores_imagenet} have the classification performance when networks are fine-tuned with an internal signal of typicality. These scores can be either normalized class predictions, or what we call `` internal probability of typicality" as it is in the first row; Or ``internal entropy of class distribution" in the second row. The last experiment (row) follows a hybrid approach, which in the first epoch samples are weighted with atypicality scores (from one-class SVM), and starting the second epoch, samples are weighted with internal scores. 

\subparagraph{Experiment with fine-tuning on PASCAL:}
In Table~\ref{tab:scores_PASCAL} we recompile previous experiments when networks were fine-tuned on ``Train Set II" (PASCAL images).  These results verify our hypothesis that we can help the generalization of CNN with weighting training examples based on functions of typicality scores. Interestingly, the performance gain after the first epoch is higher when fine-tuned on PASCAL, rather than ImageNet . We relate this to the bigger diversity in visual appearance on ImageNet collection. 

\begin{table}
	\begin{center}
		\scalebox{0.8}{\begin{tabular}{|c|c|c|c|c|c|}
				\hline
				Layers & Image & \multicolumn{4}{c|}{Weighting Functions}\\
				changed in &Set Used&\multicolumn{4}{c|}{Used in Fine-Tuning }\\
				\cline{3-6}
				fine-tuning & in Test & Atyp & Typ & Log-T& Ploy2 \\
				\hline \hline
				Top 2 FC &  Atypical & 68.17 & 64.69 & 67.41 &  69.97\\
				Top 3 FC & Atypical &  66.13 &  51.28 &  68.37 & 69.33 \\
				\hline			
				Top 2 FC &   Typical &  81.19 &  79.52 & 80.6 & 82 \\
				Top 3FC &  Typical &  78.51 & 77.1 &  76.13 & 79.41\\
				\hline			
			\end{tabular}}
		\end{center}
		\vspace*{-10pt}
		\caption{\textit{Evaluation of the effect of depth for generalization. Comparison of two alternative models, when we go further than the first fully connected layer. One with changing top two and the other one with fine-tuning top three fully connected layers. Models are fine-tuned with ``Train Set I" and MS-Hinge loss is used.}}
		\label{tab:depth}
		\vspace*{-10pt}
	\end{table}
\subsection{Investigation of The Effect of Depth}
We investigated the importance of fine-tuning deeper layers of CNN, to train models with a better generalization capacity. Table~\ref{tab:depth} shows the results of fine-tuning top-two or top-three fully connected layers of AlexNet. In the first row of each box, we changed the FC7 to have 2048 nodes. Similarly in the second row of each box, we halved the number of nodes in both FC6 and FC7. In all three models (including one reported in previous sections), we used MS-hinge loss to learn the parameters of the network. These experiments show that going deeper can hurt the fine-tuned network, especially when tested on atypical images. We would partially relate this to the limited number of images that are available for fine-tuning, therefore the network overfits to the training date. Digging deeper into this experiment with more training examples is considered as the future work. 

\section{Conclusion}
\vspace*{-5pt}
There are several points that we can conclude from this study. Atypicality is not necessary equivalent to  samples on the boundary, which typical loss functions try to emphasize in learning. The main result of this paper is that involving information about the typicality/atypicality of training samples as a weighting term in the loss function helps greatly in enhancing the performance on unseen atypical examples when training only using typical examples. We propose different ways to achieve this weighting of samples based on external (from the sample distribution) and internal signals to the network. We also found that symmetrically weighting highly typical and highly atypical examples in training gives better generalization performance. We believe that this is because the typicality/atypicality scoring of the data include information about the distribution of the samples, and therefore it incorporates in generative ``hints" to the discriminative classifier. The typicality weighting not only helps the generalization, but also helps faster learning where the network was shown to converge to significantly better results after a single epoch. 


\small
\bibliographystyle{named}
\bibliography{egbib}

\end{document}